\documentclass[]{spie}  

\usepackage{amsmath,amsfonts,amssymb}
\usepackage{graphicx}
\usepackage[colorlinks=true, allcolors=blue]{hyperref}
\usepackage{subfig}

\title{Improving Myocardium Segmentation in Cardiac CT Angiography using Spectral Information}

\author[a]{Steffen Bruns}
\author[a]{Jelmer M. Wolterink}
\author[b]{Robbert W. van Hamersvelt}
\author[a]{Majd Zreik}
\author[b]{Tim~Leiner}
\author[a]{Ivana I\v{s}gum}

\affil[a]{Image Sciences Institute, University Medical Center Utrecht, Utrecht, The Netherlands}
\affil[b]{Department of Radiology, University Medical Center Utrecht, Utrecht, The Netherlands}

\authorinfo{Further author information: (Send correspondence to S. Bruns)\\S. Bruns.: E-mail: s.bruns@umcutrecht.nl, Telephone: +31 88 75 50565}

\begin{document} 
\maketitle
\begin{abstract}
Accurate segmentation of the left ventricle myocardium in cardiac CT angiography~(CCTA) is essential for e.g. the assessment of myocardial perfusion. Automatic deep learning methods for segmentation in CCTA might suffer from differences in contrast-agent attenuation between training and test data due to non-standardized contrast administration protocols and varying cardiac output. We propose augmentation of the training data with virtual mono-energetic reconstructions from a spectral CT scanner which show different attenuation levels of the contrast agent. We compare this to an augmentation by linear scaling of all intensity values, and combine both types of augmentation. We train a 3D fully convolutional network~(FCN) with 10 conventional CCTA images and corresponding virtual mono-energetic reconstructions acquired on a spectral CT scanner, and evaluate on 40 CCTA scans acquired on a conventional CT scanner. We show that training with data augmentation using virtual mono-energetic images improves upon training with only conventional images (Dice similarity coefficient~(DSC) $0.895 \pm 0.039$ vs. $0.846 \pm 0.125$). In comparison, training with data augmentation using linear scaling improves the DSC to $0.890 \pm 0.039$. Moreover, combining the results of both augmentation methods leads to a DSC of $0.901 \pm 0.036$, showing that both augmentations lead to different local improvements of the segmentations. Our results indicate that virtual mono-energetic images improve the generalization of an FCN used for myocardium segmentation in CCTA images.
\end{abstract}


\section{INTRODUCTION}
\label{sec:intro}  
Accurate segmentation of the left ventricle~(LV) myocardium in cardiac computed tomography angiography (CCTA) images is a prerequisite for its subsequent quantitative analysis. For instance, segmentations of the myocardium can be used to assess myocardial perfusion\cite{Techasith} or to detect functionally significant coronary artery stenosis\cite{Zreik}. Manual delineation of the myocardium is time-consuming and impractical in clinical practice, especially in cases where annotations in several phases of the cardiac cycle are required. Therefore, automatic segmentation methods have been proposed. These methods mainly include atlas-based approaches\cite{Kirisli} and machine learning-based voxel classification\cite{Zreik}. Several of the latter were compared in the recent MICCAI 2017 challenge on multi-modal whole-heart segmentation (MMWHS\footnote{http://www.sdspeople.fudan.edu.cn/zhuangxiahai/0/mmwhs/}).\cite{Mortazi,Payer,Tong,Wang,Yangxin}

In order to generalize to new and unseen datasets, machine learning-based methods for CCTA segmentation require that training and test images have similar intensity histograms. Although Hounsfield units~(HU) in CT are normalized, they can differ substantially from one dataset to another in areas perfused with contrast agent. This can be due to differences in the amount of contrast agent, the flow rate, and the time between administration and acquisition. Furthermore, heart rate, cardiac output, and pathology can lead to differences in contrast enhancement of images. Generalization could be improved by including training images with a range of different HU values for contrast agent. These can be obtained by transformations on already reconstructed images, but also by inclusion of CT images that have been acquired at different x-ray energy levels. CT image acquisition with spectral CT scanners allows for the reconstruction of virtual mono-energetic CT images, which have been previously shown to provide valuable information for segmentation in CT~\cite{Chen}. In this work, we investigate whether virtual mono-energetic CT images can be used to improve the generalization of a 3D deep learning segmentation method.

\section{DATA}
\label{sec:data}
The dataset used in this study consists of 10 CCTA scans acquired on a spectral CT scanner and 40 CCTA scans acquired on a conventional CT scanner. The spectral scans were acquired on a Philips IQon CT scanner (Philips Healthcare, Best, The Netherlands). Images were acquired using a tube voltage of 120~kVp and a tube current of 120 mAs. For each acquisition, a conventional CCTA image was reconstructed, i.e. an image reconstructed from the raw data of the whole x-ray spectrum. Furthermore, virtual mono-energetic images at 60 and 90 keV were reconstructed from the same acquisition. All images were reconstructed to 0.34--0.43~$\text{mm}^2$ in-plane resolution and 0.90 and 0.45~mm slice thickness and increment, respectively. The conventional scans were acquired with a Philips Brilliance iCT scanner (Philips Healthcare, Best, The Netherlands). A tube voltage of 120~kVp and a weight-dependent tube current between 210 and 300~mAs was used. All images were reconstructed to 0.38--0.56~$\text{mm}^2$ in-plane resolution and 0.90 and 0.45~mm slice thickness and increment, respectively.

Images acquired on the spectral CT scanner were used as a training set. Reference segmentations of seven cardiac substructures were obtained in each conventionally reconstructed image. These were: LV myocardium, LV blood cavity, right ventricle blood cavity, left atrium, right atrium, ascending aorta, and pulmonary artery. To obtain a reference segmentation in each image, an initial automatic segmentation was obtained based on the MMWHS training set. These segmentations were then manually corrected by voxel painting. As all reconstructed images for a patient were based on the same acquisition and due to the dual-layer design of the scanner, we assumed perfect voxel-wise correspondence between the conventional and virtual mono-energetic images. Therefore, reference segmentations obtained in the conventionally reconstructed images were directly propagated to the corresponding virtual mono-energetic reconstructions. The 40 images obtained on a conventional CT scanner were used as a test set. Unlike in the images acquired with the spectral CT scanner, in each of these images, only the LV myocardium was manually annotated in the short-axis view in a previous study~\cite{Zreik}.

\section{METHOD}
\label{sec:method}
We trained a 3D fully convolutional network (FCN) architecture to label each voxel in an input image as one of the eight classes. The network architecture is a 3D implementation of the encoder-decoder FCN proposed by Johnson et al.~\cite{Johnson}. It consists of two downsampling layers with strided convolutions, six residual blocks with padded convolutions, and two upsampling layers with transposed convolutions. After each convolutional layer, batch normalization and rectified linear units~(ReLU) are applied.

The FCN was trained using mini-batches consisting of eight 128 $\times$ 128 $\times$ 128 patches that were randomly sampled from the training images. For all experiments, the network was trained in 13000 iterations with Adam as the optimizer. The number of classes minus the sum of soft Dice scores over all classes was used as the loss function. An initial learning rate of 0.001 was used, which was reduced by 70\% every 4000 iterations. Prior to training or testing, images were resampled to an isotropic resolution of 0.8 $\times$ 0.8 $\times$ 0.8~$\text{mm}^3$ and image intensities between -1024 and 3071 were linearly scaled between 0 and 1.

During testing, the network processed full images. The output multi-class probability maps of the network were resampled to the original resolution and each voxel was assigned the class with the highest probability. Given that most clinical scans are acquired using conventional CT scanners that do not provide spectral information, evaluation was performed using the test set containing 40 CCTA images acquired on a conventional CT scanner. This set only contained reference segmentations for the LV myocardium, hence the segmentation performance was assessed using this anatomical structure only. We evaluated agreement between the reference segmentations and automatic segmentations using the Dice similarity coefficient~(DSC) and the average symmetrical surface distance~(ASSD). The latter was calculated between the largest connected component in the automatic segmentation and the reference segmentation.

\section{EXPERIMENTS AND RESULTS}
\label{sec:experiments}
We investigated two data augmentation strategies for the training data. In the first strategy, we linearly scaled all HU values in the conventionally reconstructed images with a factor 0.5 or 2.0, and for each patient added these altered images with their corresponding segmentations to the training set. In the second strategy, we added the two virtual mono-energetic spectral CT reconstructions of each patient with their corresponding segmentations to the training set. To evaluate the efficacy of these strategies, four experiments were performed. In Experiment~1, we trained an FCN using only the conventional reconstructions. In Experiment~2, data augmentation by intensity-based scaling was used. In Experiment~3, data augmentation by inclusion of two virtual mono-energetic images was used. Finally, in Experiment~4, results of both strategies were combined and evaluated.

\begin{figure}[tb]
	\centering
    {
  		\includegraphics[width=0.8\textwidth]{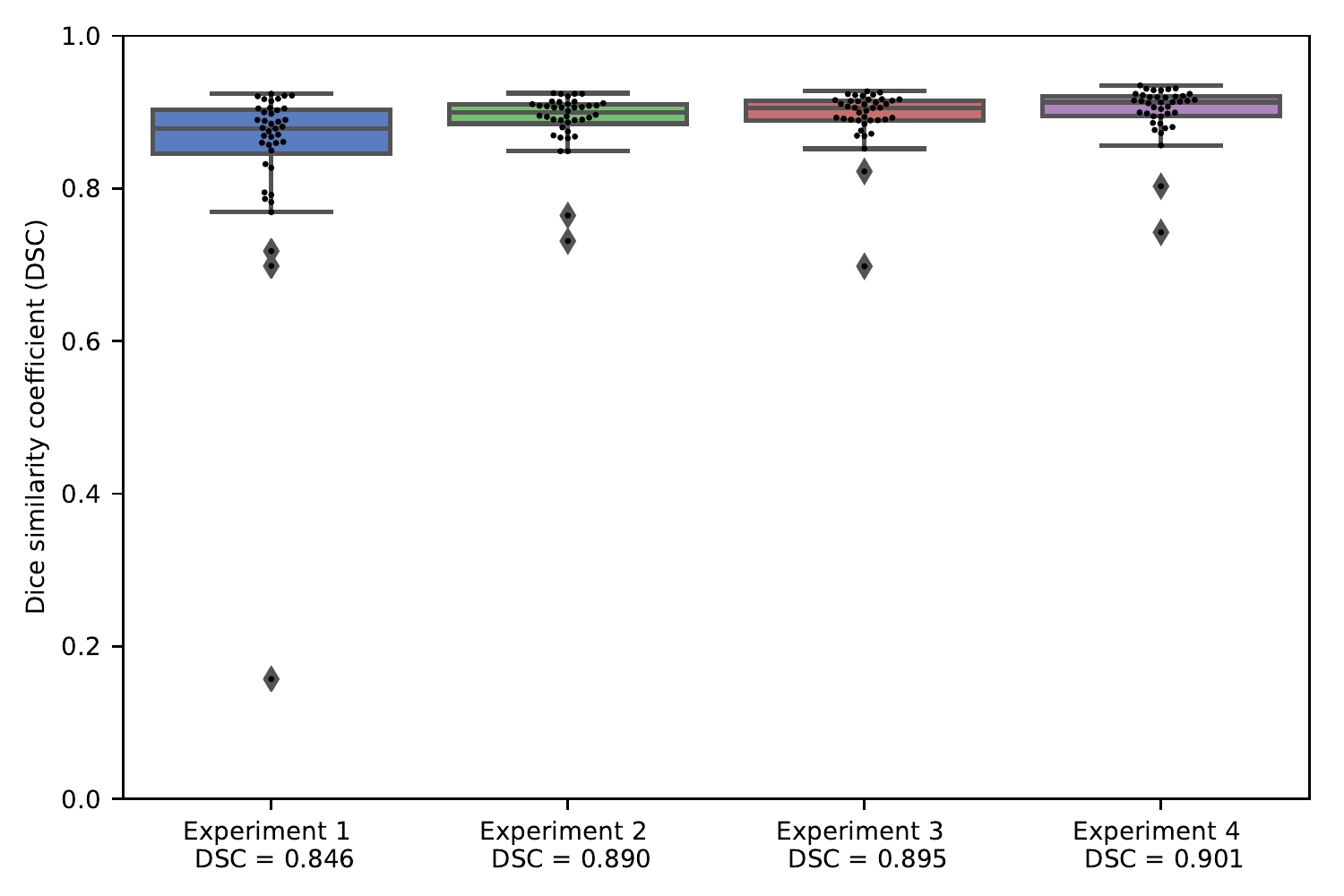}
    }
   	\caption[Dices]
    {\label{fig:Dices} Box plots showing the DSC between automatic and reference segmentations in 40 conventional CCTA test images. Results are shown for (Experiment~1) an FCN trained on only conventional images, (Experiment~2) an FCN trained on conventional images and linearly scaled conventional images, (Experiment~3) an FCN trained on conventional images and two different virtual mono-energetic images per patient, (Experiment~4) the combination of Experiments 2 and 3.}
\end{figure}

\begin{figure}[tb]
	\centering
    {
  		\includegraphics[width=0.8\textwidth]{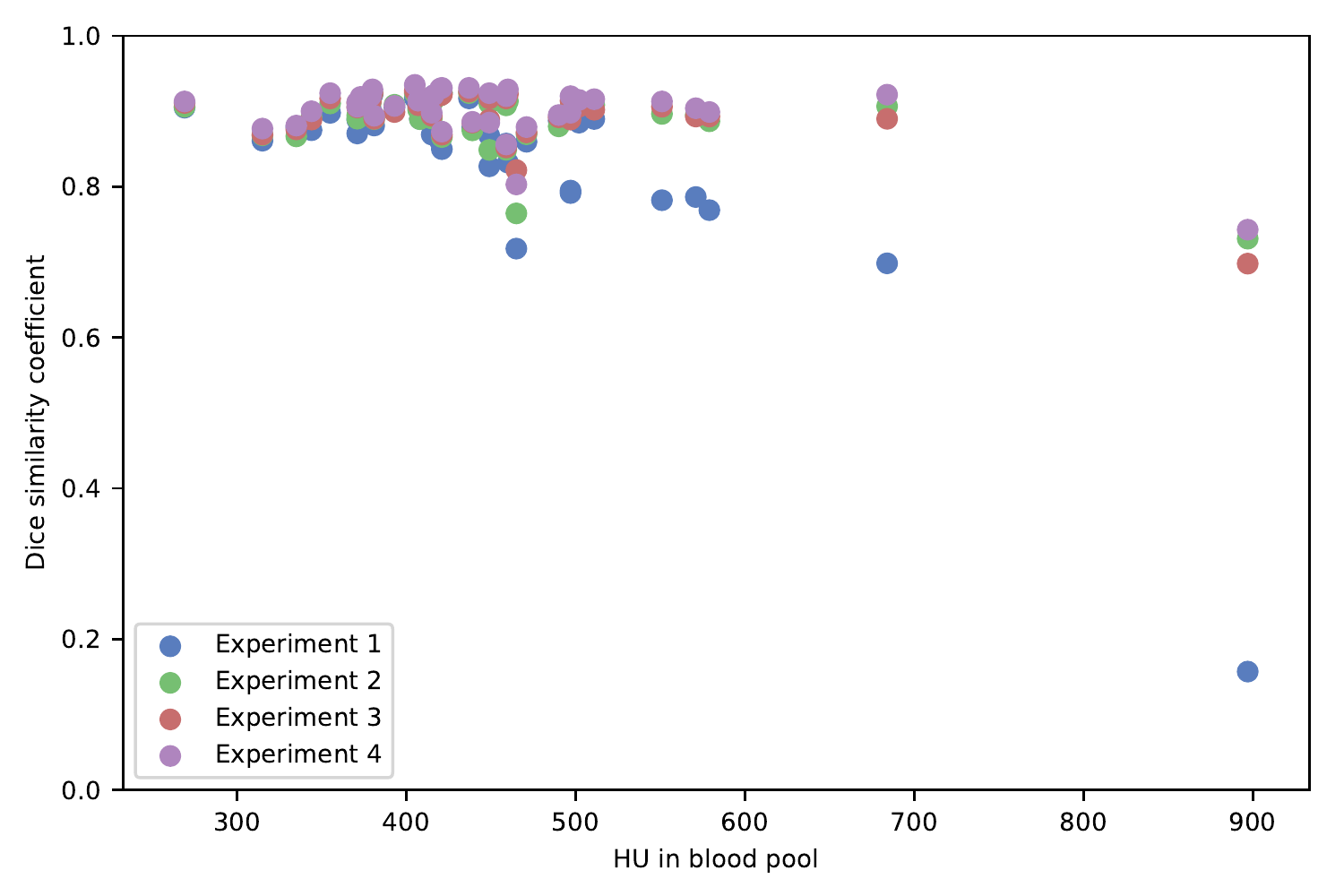}
    }
    \caption[HU]
    {\label{fig:HU} Scatter plot showing the DSC between automatic and reference segmentations in 40 conventional CCTA test images. Results are shown for (Experiment~1) an FCN trained on only conventional images, (Experiment~2) an FCN trained on conventional images and linearly scaled conventional images, (Experiment~3) an FCN trained on conventional images and two different virtual mono-energetic images per patient, (Experiment~4) the combination of Experiments 2 and 3. Results for the four experiments on all test images are shown over the average HU in the blood pool.}
\end{figure}

To reflect application to typical CCTA acquisitions, which usually do not include virtual mono-energetic images, we evaluated our experiments on images obtained on a conventional CT scanner. Figure~\ref{fig:Dices} shows box plots for the DSC obtained on the 40 test scans in all four experiments. 
Training using only conventional images (Experiment 1) resulted in an average DSC of $0.846 \pm 0.125$ and an ASSD of $2.03 \pm 2.77$~mm. The intensity-based data augmentation (Experiment 2) led to an average DSC of $0.890 \pm 0.039$ and an ASSD of $1.42 \pm 0.40$~mm. Augmenting the dataset with two virtual mono-energetic images (Experiment 3) increased the average DSC to $0.895 \pm 0.039$ and decreased the ASSD to $1.38 \pm 0.48$~mm. Finally, combining the results of both augmentations (Experiment 4) led to a further increase of the average DSC to $0.901 \pm 0.036$ and an ASSD of $1.38 \pm 0.43$~mm. The Friedman test and pair-wise Wilcoxon signed-rank tests with Bonferroni correction were used to validate that the resulting DSC for all four experiments were significantly different from each other (all p-values $< 0.025$). The largest improvements in DSC were gained for images with high HU values in the blood pool as can be seen in Fig.~\ref{fig:HU}.
The conventional training images only had average HU values between 271 and 343 in the blood pool. Therefore, test cases in which the intensities in the blood pool were substantially higher could benefit the most from augmentation. Although this figure shows a clear relation between contrast-agent attenuation, data augmentation, and segmentation performance, some issues like anatomical abnormalities could not be solved by these data augmentation approaches.

\begin{figure}[tb]
   \begin{center}
   \includegraphics[width=0.65\textwidth]{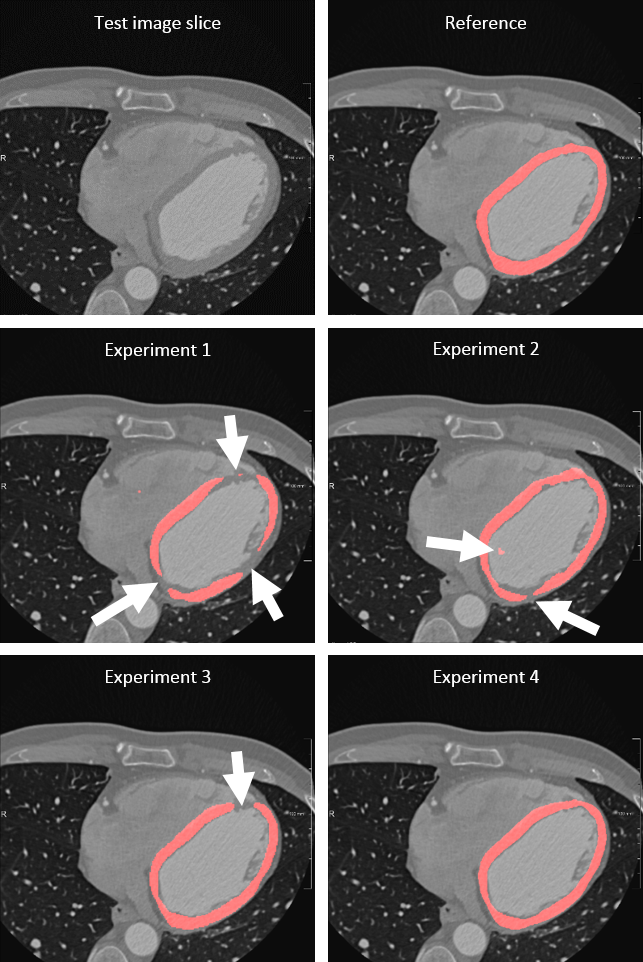}
   \end{center}
   \caption[Segmentations] 
   { \label{fig:Segmentations} Example segmentation results. The first row shows the image and the reference segmentation. The second and third row show results obtained in Experiments 1 to 4.}
\end{figure}

Figure~\ref{fig:Segmentations} shows segmentation results for a test image where augmentation improved the segmentation performance. 
The reference segmentation is shown in the upper right, the segmentations obtained by the four experiments are shown in the subsequent rows. The network trained only on conventional images failed to segment the myocardium in several areas, whereas both types of augmentation resulted in an improved segmentation of the myocardium. However, both methods improved the segmentation of different parts of the myocardium. By combining both predictions, a segmentation closely matching the reference could be obtained. Although both augmentations led to similar improvements in DSC, it is likely that different data augmentations have different local effects.

\section{NEW OR BREAK-THROUGH WORK TO BE PRESENTED}
This work shows that virtual mono-energetic CCTA reconstructions that are standardly acquired along with conventional CCTA reconstructions on a spectral CT scanner can be used as data augmentation for the segmentation of cardiac structures.

\section{CONCLUSION}
The results indicate that using virtual mono-energetic image reconstructions as data augmentation can improve the generalization capabilities of an FCN trained for cardiac segmentation in CCTA. Combining these results with linear intensity scaling augmentation can further improve generalization.

\acknowledgments 
This work is part of the research programme Deep Learning for Medical Image Analysis with programme number P15-26, which is financially supported by the Netherlands Organisation for Scientific Research (NWO) and Philips Healthcare. We gratefully acknowledge the support of NVIDIA Corporation with the donation of the Titan Xp GPU used for this research.\\

\bibliography{report} 

\begin{thebibliography}{10}

\bibitem{Techasith}
Techasith, T. and Cury, R.~C., ``Stress myocardial {CT} perfusion: An update
  and future perspective,'' {\em JACC Cardiovascular Imaging}~{\bf 4},
  905--916 (2011).

\bibitem{Zreik}
Zreik, M., Lessmann, N., van Hamersvelt, R.~W., Wolterink, J.~M., Voskuil, M.,
  Viergever, M.~A., Leiner, T., and I\v{s}gum, I., ``Deep learning analysis of
  the myocardium in coronary {CT} angiography for identification of patients
  with functionally significant coronary artery stenosis,'' {\em Medical Image
  Analysis}~{\bf 44},  72--85 (2018).

\bibitem{Kirisli}
Kirisli, H.~A., Schaap, M., Klein, S., Papadopoulou, S.~L., Bonardi, M., Chen,
  C.~H., Weustink, A.~C., Mollet, N.~R., Vonken, E.~J., van~der Geest, R.~J.,
  van Walsum, T., and Niessen, W.~J., ``Evaluation of a multi-atlas based
  method for segmentation of cardiac {CTA} data: a large-scale, multicenter,
  and multivendor study,'' {\em Medical Physics}~{\bf 37},  6279--6291 (2010).

\bibitem{Mortazi}
Mortazi, A., Burt, J., and Bagci, U., ``Multi-planar deep segmentation networks
  for cardiac substructures from {MRI} and {CT},'' in [{\em Statistical Atlases
  and Computational Models of the Heart}{\nolinebreak\hspace{0.1em}]},  Pop,
  M., ed., {\em Proc. STACOM} {\bf 10663},  199--206 (2017).

\bibitem{Payer}
Payer, C., \v{S}tern, D., Bischof, H., and Urschler, M., ``Multi-label whole
  heart segmentation using {CNN}s and anatomical label configurations,'' in
  [{\em Statistical Atlases and Computational Models of the
  Heart}{\nolinebreak\hspace{0.1em}]},  Pop, M., ed., {\em Proc. STACOM} {\bf
  10663},  190--198 (2017).

\bibitem{Tong}
Tong, Q., Ning, M., , Si, W., Liao, X., and Qin, J., ``3{D} deeply-supervised
  {U}-{N}et based whole heart segmentation,'' in [{\em Statistical Atlases and
  Computational Models of the Heart}{\nolinebreak\hspace{0.1em}]},  Pop, M.,
  ed., {\em Proc. STACOM} {\bf 10663},  224--232 (2017).

\bibitem{Wang}
Wang, C. and Smedby, {\"O}., ``Automatic whole heart segmentation using deep
  learning and shape context,'' in [{\em Statistical Atlases and Computational
  Models of the Heart}{\nolinebreak\hspace{0.1em}]},  Pop, M., ed., {\em Proc.
  STACOM} {\bf 10663},  242--249 (2017).

\bibitem{Yangxin}
Yang, X., Bian, C., Yu, L., Ni, D., and Heng, P., ``Hybrid loss guided
  convolutional networks for whole heart parsing,'' in [{\em Statistical
  Atlases and Computational Models of the Heart}{\nolinebreak\hspace{0.1em}]},
  Pop, M., ed., {\em Proc. STACOM} {\bf 10663},  215--223 (2017).

\bibitem{Chen}
Chen, S., Zhong, X., Hu, S., Dorn, S., Kachelrie{\ss}, M., Lell, M., and Maier,
  A., ``Automatic multi-organ segmentation in dual energy {CT} using 3{D} fully
  convolutional network,'' in [{\em Medical Imaging in Deep
  Learning}{\nolinebreak\hspace{0.1em}]},  (2018).

\bibitem{Johnson}
Johnson, J., Alahi, A., and Fei-Fei, L., ``Perceptual losses for real-time
  style transfer and super-resolution,'' in [{\em European Conference on
  Computer Vision}{\nolinebreak\hspace{0.1em}]},   694--711 (2016).

\end{thebibliography}
\bibliographystyle{spiebib} 

\end{document}